\pdfoutput=1

\documentclass[11pt]{article}

\usepackage{ACL2023}
\usepackage{times}
\usepackage{latexsym}
\usepackage[T1]{fontenc}
\usepackage[utf8]{inputenc}
\usepackage{microtype}
\usepackage{inconsolata}
\usepackage{fancyhdr}

\usepackage[frozencache=true,cachedir=minted-cache]{minted}
\usepackage{graphicx}
\usepackage{enumitem}
\usepackage{float}
\usepackage{xcolor}

\definecolor{prompt_color}{HTML}{EEEEEE}
\newcommand{\bc}[1]{\colorbox{prompt_color}{#1}}

\title{Alfred: A System for Prompted Weak Supervision}

\author{Peilin Yu \qquad
  Stephen H. Bach \\
  Department of Computer Science \\
  Brown University \\
  \texttt{\{peilin\_yu, stephen\_bach\}@brown.edu} \\}

\begin{document}

    \maketitle
    
    \begin{abstract}
Alfred is the first system for programmatic weak supervision (PWS) that creates training data for machine learning by prompting.
In contrast to typical PWS systems where weak supervision sources are programs coded by experts, Alfred enables users to encode their subject matter expertise via natural language prompts for language and vision-language models.
Alfred provides a simple Python interface for the key steps of this emerging paradigm, with a high-throughput backend for large-scale data labeling.
Users can quickly create, evaluate, and refine their prompt-based weak supervision sources; map the results to weak labels; and resolve their disagreements with a label model.
Alfred enables a seamless local development experience backed by models served from self-managed computing clusters.
It automatically optimizes the execution of prompts with optimized batching mechanisms. 
We find that this optimization improves query throughput by 2.9$\times$ versus a naive approach. 
We present two example use cases demonstrating Alfred on YouTube comment spam detection and pet breeds classification.
Alfred is open source, available at \href{https://github.com/BatsResearch/alfred}{https://github.com/BatsResearch/alfred}.
\end{abstract}
    \section{Introduction}
Acquiring labeled data is a significant challenge for machine learning for its time-consuming and expensive nature. 
Programmatic weak supervision (PWS) provides a more efficient method of data annotation by using noisy heuristics to label data. 
In a typical PWS setup, domain experts design labeling functions (LFs) as programs that vote for a label or abstain. \cite{ratner2016data, ratner2017snorkel} 
Recently, there has been a growing interest in creating LFs from large, pre-trained models through prompting~\cite{smith2022language,arora2022ask, zhang2022prboost}.
In the shift to this new setting, executing LFs goes from the least to the most computationally expensive part of the process, highlighting the importance of providing a software infrastructure that facilitates efficient development.
However, existing toolkits for large language models mainly prioritize prompt templating and tuning, leaving an unmet need for a system that that connects prompting with the creation of training data.

\begin{figure*}[t]
    \centering
    \includegraphics[width=\textwidth]{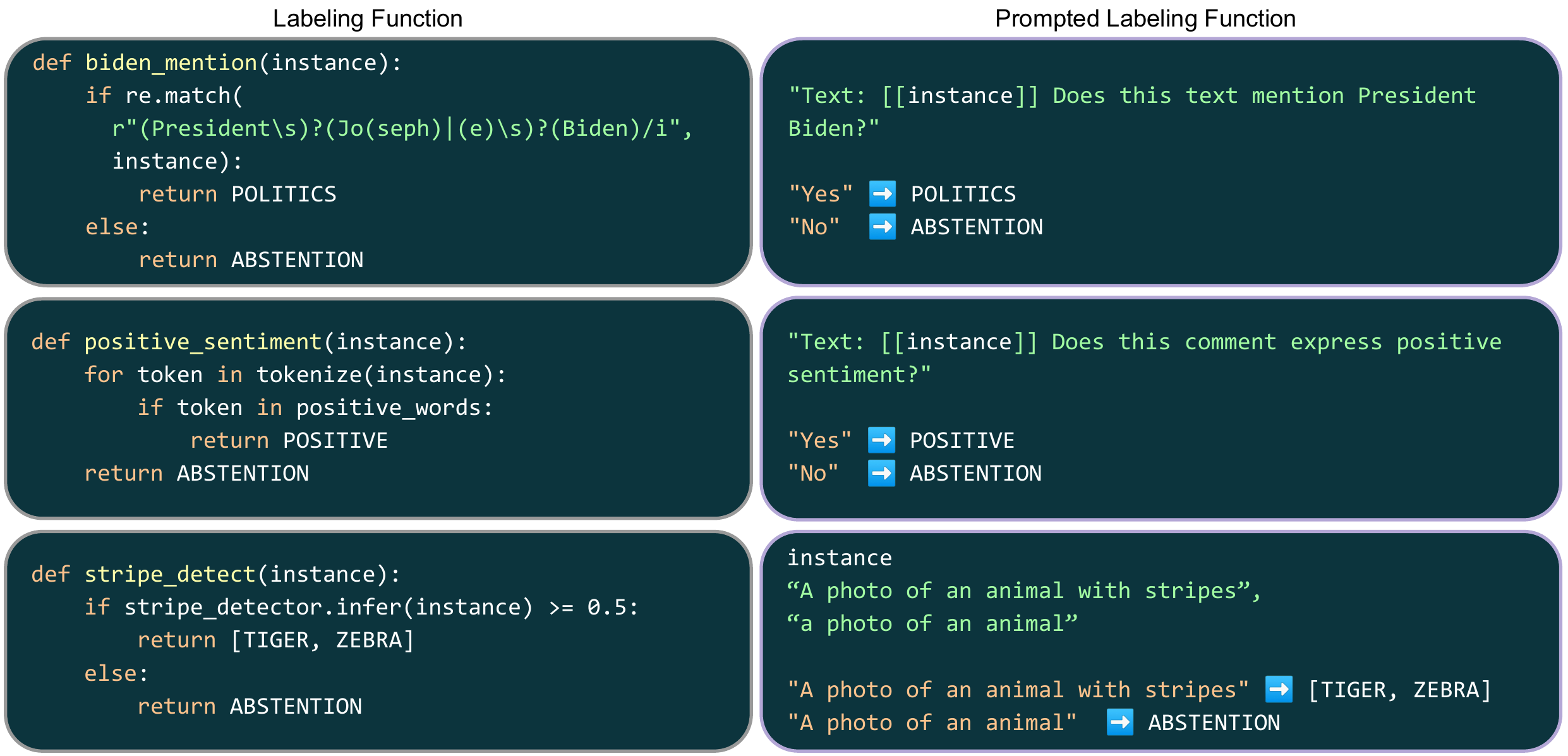}
    \caption{Examples of a labeling function versus a prompted labeling function.
    For the first example, each expresses supervision relating mentions of President Biden to the category of politics.
    Instead of specifying an intricate regular expression, a prompted labeling function uses the prompt “Text: [[instance]] Does this text mention President Biden?” where [[instance]] is replaced by the news article to be labeled.
    The response is mapped to a vote on the true label. 
    The second example demonstrates how heuristics about positive sentiment that were previously hard to define can be flexibly expressed as a natural language question.
    Instead of defining a set of keywords for fuzzy sentiment matching, we can simply ask large pretrained models for answers about the sentiment.
    For the third example, we consider an animal labeling task where we use the visual attributes ``stripes'' to vote for the classes TIGER and ZEBRA.
    Previously, we would need to first collect supervised training data for attributes like stripes and then train classifiers to make the decisions.
    With modern vision-language models (e.g. CLIP), we can simply express the attribute detection task as a set of candidate prompts.    
    }
    \label{fig:alfred_plf}
\end{figure*}

Prompted models offer a unique opportunity to enhance existing PWS systems. 
Traditional PWS systems require programming LFs with code that specifies heuristic domain knowledge.
With large pre-trained models, natural language-based prompts can be used as LFs, also known as prompted LFs \cite{smith2022language}. 
This approach allows the easy expression of complex rules that were previously difficult to specify using code, as the example in Figure~\ref{fig:alfred_plf} shows.
The ability to use prompts to define labeling functions simplifies and streamlines the weak supervision process, as well as potentially elevating the quality of the annotations.
This benefit is particularly helpful for tasks involving computer vision, where previously PWS has been limited to tasks for which models can identify key features (such as objects) over which to program rules.
Whether the domain is language or multi-modal, prompts let users experiment with different heuristics (and phrasings of those heuristics).
Therefore, enabling an iterative development experience is essential for the success of a prompt-based PWS system.

The switch to prompted models for weak supervision also presents significant challenges.
It first requires rethinking the abstractions and workflow of first-generation PWS systems.
Instead of editing code and managing libraries of functions, users must manage libraries of prompts, track their outputs on multiple datasets, and develop strategies for mapping those outputs to labels.
This change is complicated by large models’ demand for computational resources.
The throughput of the models is a new development bottleneck.
The extra overhead of remotely hosted models is a further hindrance to the iterative workflow of weak supervision.

Existing open-source software for prompting concentrates on prompt engineering \cite{orr2022manifest, bach2022promptsource}, prompt chains \cite{Chase_LangChain_2022} or continuous (i.e., soft) prompt tuning \cite{ding-etal-2022-openprompt}; placing less emphasis on throughput for a large-model-in-the-loop workflow.
Additionally, many existing open-source systems have not developed support for vision-language models, despite their benefits for data labeling.
Incorporating large pre-trained models into a PWS system remains an open problem in the open-source software space, requiring innovative solutions to address these challenges and complement existing software focused on other aspects of prompting.

We present Alfred, a versatile PWS system that leverages large pre-trained models for image and text annotations. Alfred aims to provide an environment for the rapid development of prompt-based supervision, while maintaining a consistent development experience similar to established PWS systems.
We designed Alfred with usability and efficiency in mind, aiming to provide a rapid and smooth experience for developing prompt-based supervision.
Alfred supports popular large language models from Hugging Face's transformer package \cite{wolf-etal-2020-transformers}, including the GPT family \cite{radford2019language}, the T5 family \cite{raffel2020exploring}, etc., and vision-language models like CLIP \cite{radford2021learning}, etc.
Alfred also supports local ONNX models, or API-based models from OpenAI, AI21, and Cohere.
Moreover, Alfred provides easy templating tools to help users quickly create, evaluate, and refine prompted LFs.
Alfred offers easy ways to deploy inference servers remotely, in addition to local model hosting.
Alfred also optimizes model inference throughput with improved batching techniques and provides utilities for efficient LLM deployment and interaction.
Finally, Alfred contains a library of label models to distill the outputs of prompted labeling functions into the final training datasets for downstream end models.

\begin{figure*}[t]
    \centering
    \includegraphics[width=\textwidth]{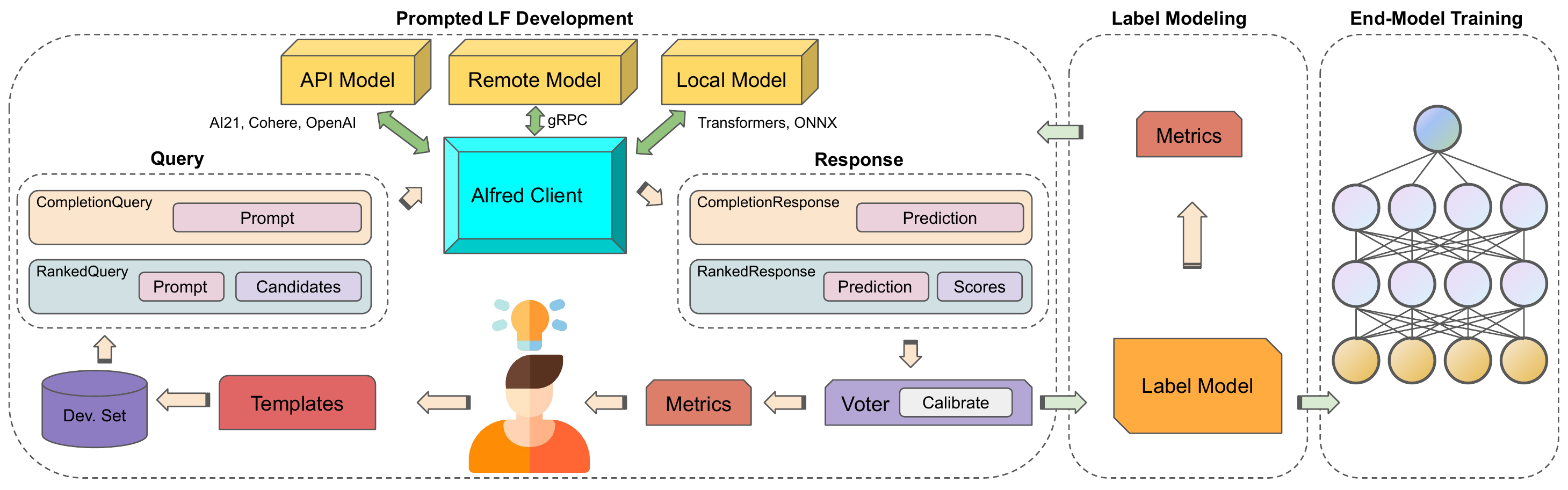}
    \caption{A typical workflow for programmatic weak supervision with Alfred. First, developers use Alfred to iteratively design, evaluate, and refine their prompted labeling functions (LFs).
    They use prompt Templates to generate Queries to Models based on data.
    The responses of Models are mapped to votes on the true labels for examples by Voters, which can be calibrated. 
    Then the included Label Models combine the noisy votes to produce probabilistic training labels for an end model.}
    \label{fig:alfred_loop}
\end{figure*}

Alfred is a prototype for a second generation of PWS systems with prompting at their core.
To this end, we highlight three key feature of Alfred:
\begin{itemize}[noitemsep,nolistsep,leftmargin=*]
\item \textbf{Prompt-based weak supervision for images and text.}
Alfred provides the necessary tools for users to create, evaluate, and refine prompt templates for both image and text weak supervision tasks. The inclusion of a query caching system that automatically stores and updates model responses facilitates development.
\item \textbf{Optimized inference throughput.}
Alfred implements a dynamic batching mechanism that optimizes large sets of prompts. This feature allows models hosted by Alfred to achieve 2-3$\times$ greater throughput than naive implementations.
\item \textbf{Seamless local development experience.}
Alfred can host models remotely and make them accessible to developers via gRPC, a high-throughput protocol for remote procedure calls.\footnote{\href{https://grpc.io/}{grpc.io}}
It enables sending datasets to servers in large chunks to maintain higher throughput.
Alfred also implements a SSH-based port-forwarding utility for the gRPC connection, easing deployment on shared clusters such as those at universities.

\end{itemize}
    \section{Related Work and Background}
Alfred sits at the intersection of programmatic weak supervision and large pretrained models. 
In this section, we overview the most related work.

\noindent
\textbf{Programmatic Weak Supervision.}
Traditionally, supervised learning relied on manually labeled data, and data labeling has been seen as a key bottleneck for many applications.
Recently, a family of programmatic weak supervision (PWS) methods have offered an alternative  to costly manual annotations by incorporating multiple sources of noisy labels to create training datasets~\cite{zhang2022survey}.
Typically, a PWS system such as Snorkel~\cite{ratner2017snorkel} has a three-stage setup: 
First, developers will create heuristics called labeling functions (LFs) that vote on the label for an example (or abstain). 
Second, a label model will reconcile the noisy votes and provide probabilistic labels for the data. 
Finally, freshly annotated data is used to train an end model with a noise-aware loss objective (e.g. in a classification setting, this can be a soft cross entropy~\cite{ratner2016data}).
Alfred focuses on the first two stages and aim to efficiently incorporate modern large pretrained models into the development workflow.

\noindent
\textbf{Prompting for Pre-Trained Models.}
With the emergence of large pre-trained langugae and vision-language models, prompting has become a popular approach to many few-shot and zero-shot tasks~\cite{brown2020gpt3, schick-schutze-2021-shot, radford2021learning}.
Prompting can create training examples, generate data, modify datasets, and improve model reasoning~\cite{schick-schutze-2021-generating, ye2022zerogen, chia2022relationprompt, wu-etal-2022-generating, wang2022self, wei2022chain, zelikman2022star}.
This presents a unique opportunity for combining prompting for large pretrained models and weak supervision.
Recent studies have investigated strategies to combine large language models into weak supervision frameworks~\cite{smith2022language,chen2022shoring, arora2022ask, zhang2022prboost}.
Alfred aims to provide a platform for the rapid development of weak supervision applications that rely on large pre-trained language and vision-language models, as well as enable experimentation with new ways of using those models.

\noindent
\textbf{Systems for Prompt Development.}
Prompting has led to the development of software toolkits that aid in prompting and research across various tasks.
Many tools have been developed for various use cases with large language models.
PromptSource~\cite{bach2022promptsource} is a development environment for creating and archiving sets of prompts.
OpenPrompt~\cite{ding-etal-2022-openprompt} is a library focused on tuning prompts and prompted models with training data.
Manifest~\cite{orr2022manifest} provides a unified front end for prompting large language models via different APIs.
LangChain~\cite{Chase_LangChain_2022} provides convenient utilities for building applications that chain together multiple prompts and outputs.
To complement the existing tools in this growing space, Alfred is designed to be a PWS system based on prompting both large language and vision-language models.
    \section{Prompted LF Development}

Alfred is designed to enable the development and application of prompted labeling functions (LFs)
Compared to the typical workflow of PWS systems, where developing LFs is not computationally demanding, developing prompted LFs has model inference as a bottleneck. 
These large models are often hosted remotely on virtual instances or computing clusters, which can add to the challenge of iterative prompt development.
In an iterative development environment, creating prompted labeling functions requires a platform that provides optimal throughput and low latency for a rapid local development experience. 
To illustrate Alfred’s key focuses, we illustrate a typical workflow (Figure~\ref{fig:alfred_loop}) for using Alfred to create a training dataset:

\noindent
\textbf{Step 1: Task Setup} For a large model to be used in the development loop, developers can elect to use either self-hosted models or API-based models. 
For self-hosted models, Alfred provides an \bc{AlfredServer} to host the model on cloud or cluster nodes.
As the main development interface, the user can simply start a \bc{Client} by specifying the type of model.
Before creating prompted LFs, users need to familiarize themselves with the task by exploring the raw, unlabeled dataset. 
If there is no development subset available, the developer can annotate a small portion of the data as a held-out evaluation benchmark. 
Alfred implements a \bc{Dataset} class based on Apache Arrow for fast data access.
User may load a \bc{Dataset} from CSV, JSON or Dataframe objects.
It also offers direct support for datastes from the Hugging Face `Dataset` package \cite{lhoest2021datasets}.
During the exploration process, developers may gain insights into the data and the label space, and identify potentially useful heuristics. 
Moreover, users can freely experiment with prompts with a few unlabeled instances by directly interacting with the \bc{Client}.

\noindent
\textbf{Step 2: Iterative Prompt Development}: When the user is ready for prompt development, they can use a \bc{Template} to define a prompted LF for either text completion or scoring schemes.
A \bc{Template}, given a \bc{Dataset}, will produce the corresponding \bc{Query} objects.
\bc{Client} will return the appropriate \bc{Response} object for each \bc{Query}.
To map the model responses to votes, users define the corresponding \bc{Voter} and identify the label maps and matching functions to be used for each prompted LFs.
Label maps define how potential model responses are associated with the label space. 
Matching functions specify how the \bc{Voter} determines a match. 
By default, Alfred employs an exact match mechanism, but this can be substituted with user-defined matching functions for uncased matching or  embedding similarity matching, etc.
A \bc{Voter} can be optionally calibrated to reduce model-specific biases~\cite{zhao2021calibrate}.
With the model responses and the Voter, users can obtain the label votes for each of their prompted LFs and examples in a matrix format.
Finally, users can evaluate the quality of their prompted LFs using a set-aside development \bc{Dataset} with desired metrics.
Here, users can continue to refine their prompted LFs and iterate as necessary.
Once users are satisfied with the performance of their development benchmark, they may proceed.

\noindent
\textbf{Step 3:  Aggregate Prompt Responses} 
Finally, Alfred can aggregate the votes from each \bc{Voter} with a \bc{LabelModel} to produce probabilistic estimates of the true labels for the examples.
Alfred also supports \emph{partial labels}, i.e., labels that narrow down the possible set of classes but are not specific enough to vote on a single class~\cite{yu2022nplm}.
The probabilistic labels can then be used to train a wide range of end models.
    
\section{System Design}
In this section, we describe and highlight the key design decisions for Alfred. 

\subsection{Query and Response Types}

\begin{figure}[t]
    \centering
    \begin{minted}[mathescape,
               numbersep=4pt,
               gobble=2,
               frame=lines,
               fontsize=\small,
               framesep=0mm]{python} 
               
    from alfred import Client
    from alfred.fm.query import RankedQuery, 
                                CompletionQuery
    LMClient = Client(...)

    headline = "Liverpool wins 7-0!"
    
    LMClient(
        CompletionQuery(headline
        + " What is the topic of this headline?")
    )
    # Example Response:
    # >> CompletionResponse(prediction="Football")
    
    LMClient(
        RankedQuery(headline
        + " What is the topic of this headline?",
          candidate=["Sports", "Politics",
                     "Tech", "Business"])
    ) 
    # Example Response:
    # >> RankedResponse(prediction="Sports", 
    #      scores={"Sports":0.76, "Politics":0.10,
    #              "Tech":0.07, "Business":0.07 })
        
    \end{minted}
    \caption{Typed \bc{Query} and \bc{Response} in Alfred}
    \label{fig:alfred_client_code}
\end{figure}

We identify two main patterns using prompts for PWS: text completion and scoring. 
Text completion is when a language model generates responses using a heuristic decoding strategy over the whole model vocabulary, while scoring is when a language ranks candidate completions or a vision-language model ranks candidate prompts, i.e., captions.
Alfred implements typed \bc{Query} and \bc{Response} classes for these two patterns (Figure~\ref{fig:alfred_client_code}).
Upon applying the \bc{Template} operation on a dataset instance, it produces either a \bc{CompletionQuery} or a \bc{RankedQuery} for each instance based on the \bc{Template} definition.
 The resulting query can be directly fed into the \bc{Client}. The \bc{Client} then returns a corresponding \bc{CompletionResponse} or \bc{RankedResponse} with the prediction as the main payload, along with any other requested or useful information, such as the logits for each candidate.

\subsection{Templates for Prompted LFs}
Prompt templates are at the core of systems for prompting.
In Alfred, prompt templates are expressed as \bc{Template} objects.
For natural language tasks, users use the \bc{StringTemplate} class.
To produce \bc{Query} objects, users can call `Template.apply(instance)' or `Template.apply\_to\_dataset(dataset).'
A \bc{StringTemplate} is defined with a template string with keywords enclosed by double square brackets, e.g. ``[[text]] Does the previous context express spouse relation between [[entity\_a]] and [[entity\_b]]?''.
An optional field for \bc{Template} is `answer\_choices,' where one may specify the candidate completions.
By specifying the `answer\_choices,' the  \bc{StringTemplate} would yield a \bc{RankedQuery}.
An example code snippet showing the creation of a \bc{RankedQuery} is in Figure~\ref{fig:alfred_template_code}.
For image annotation tasks, users may define an \bc{ImageTemplate} by specifying the candidate prompts.
Upon applying to images, \bc{ImageTemplate} will produce \bc{RankedQuery} objects with the images and candidate prompts.

\begin{figure*}[t]
    \centering
    \begin{minted}[mathescape,
               numbersep=5pt,
               gobble=2,
               frame=lines,
               fontsize=\small,
               framesep=1mm]{python}
    from alfred.template import StringTemplate, ImageTemplate
    
    string_template = StringTemplate(
        "Context: [[text]]\n\nIs the above text about weather?", answer_choices = ["Yes", "No"]
    )
    example = {'text': "A pleasant day with a sunny sky."}
    prompt = string_template.apply(example) 
    # >> RankedQuery("Context: A pleasant day with a sunny sky.\n\nIs the above text about weather?",
    #                candidates=["Yes", "No"])

    image_template = ImageTemplate(
        {"label": ["cat", "dog"]},
        template = "A photo of [[label]]."
    )
    example = cat_image
    prompt = image_template.apply(example)
    # >> RankedQuery(example, candidates=["A photo of cat.", "A photo of dog."])
    \end{minted}
    \vspace{-1ex}
    \caption{Example code snippet for creating a \bc{RankedQuery} from a \bc{StringTemplate} or a \bc{ImageTemplate}.}
    \label{fig:alfred_template_code}
    \vspace{-1ex}
\end{figure*}

\subsection{Throughput Optimization}
Alfred is designed to handle large numbers of queries.
Self-hosted models from the Transformers package~\cite{wolf-etal-2020-transformers} are set up to use model parallelization enabled by Accelerate \cite{accelerate}, with user-customizable device maps for parallelizing the model across multiple GPUs.
Alfred adopts a dynamic batching strategy that groups instances with similar lengths together and adjusts the input batch size dynamically to maximize model inference throughput.
The core idea of the dynamic batching strategy is to group input instances with similar token lengths to minimize padding and maximize memory utilization.

With the dynamic batching strategy, on a node with 8 NVIDIA Tesla V100s, Alfred achieves a speedup of up to 2.5$\times$ and a token throughput increase of 2.9$\times$ for approximately 500 queries ($\sim$21,000 tokens) compared to an unoptimized strategy with T0++~\cite{sanh2022multitask}, an 11-billion parameter T5-based \cite{raffel2020exploring} language model in FP32.
Additionally, Alfred includes a client-side query caching system that automatically stores and updates model responses to facilitate prompt development and avoid redundant queries during development.
Alfred also implements a server-side caching system for large multi-modal pretrained models such as CLIP.
At inference time, Alfred will cache the input data with its corresponding encoded latent representations from different encoder head for each modalities.
This server-side caching system effectively avoids redundant encoding computation on the server end.

\subsection{Remote Self-Hosting of Models}
The computational demands of large pre-trained models can pose a challenge when using them for weak supervision development.
To address this challenge, Alfred provides utilities for deploying and interacting with models on  remote virtual instances or computing clusters. 
Additionally, Alfred implements a SSH-based tunneling service that ensures a secure local connection while preserving all Alfred functionality.
The tunneling utility also simplifies deployment of the server on shared computing clusters, with the login node serving as a jump host for the computing node. 
This is particularly useful for using Alfred on centrally-managed shared computing clusters such as those at universities.
Alfred's built-in SSH tunneling is also capable of handling 2-factor authentication, which is common for shared clusters.
To enable efficient communication between the client and server, Alfred uses gRPC, a high-performance, open-source remote procedure call framework.
This enables Alfred to provide a seamless development experience for weak supervision development without the need for expensive local resources.

\subsection{Mapping Responses to Votes}

Another core piece of the Alfred system is the \bc{Voter} class.
Each \bc{Voter} defines how to map model responses to votes for the true label of an exmaple.
The votes can be class labels or partial labels (e.g., attributes) specified by the users.
The voting mechanism also relies on a matching function, which by default only casts a vote for an exact match. 
Users may provide their intended matching mechanisms such as uncased matching or embedding similarity matching for more flexibility for each \bc{Voter}.
Furthermore, \bc{Voter} can be contextually calibrated for the specific \bc{Template} class to reduce model bias towards predicting certain answers.
Recent studies show calibration can be helpful for many prompt-based tasks~\cite{zhao2021calibrate, smith2022language}.
By calling `Client.calibrate(Template, Voter),' Alfred will calibrate the voting weights according to the strategy proposed by \citeauthor{zhao2021calibrate} and automatically apply the calibration during voting.

\subsection{Label Models for Aggregating Votes}

Alfred currently includes four label models for combining the votes from prompted labeling functions.
The four label models are available to meet different use cases.
The \bc{MajorityVote} model is a baseline option suitable for fast development iteration, while the \bc{NaiveBayes} model is recommended as the standard label model.
Alfred also includes \bc{NPLM}~\cite{yu2022nplm} (noisy partial label model) to support weak supervision from partial labels, which are labels that narrow down the possible set of classes but are not specific enough to vote on a single class.
\bc{FlyingSquid} \cite{fu2020fast} is the fourth model option and is recommended when \bc{MajorityVote} is not accurate enough but more speed than \bc{NaiveBayes} is needed.
These label model classes have a unified interface, providing a consistent experience for users.
After processing votes, the label model module generates probabilistic labels, represented as a distribution over the label space, for the given unlabeled dataset.
Finally users can use the estimated probabilistic labels to train an end model for the downstream task.
    \section{Example Use Cases}

In this section, we present two example use cases for how Alfred can be used to create training data for specific machine learning tasks using natural language prompts in both text and image domains.
We measure the labeling quality by taking the top-1 accuracy of the estimated probabilistic labels given by the label model.
The notebooks to reproduce these examples are in the Alfred repository.

\subsection{Youtube Comment Spam Detection}
\begin{table}[H]
    \centering
    \begin{tabular}[width=0.5\textwidth]{ccc}
    \hline
     Zero Shot &  Prompted LFs & Prompted LFs+C\\
         \hline
          46.8 & 57.8 & 65.3\\
         \hline
     \end{tabular}
    \caption{Top-1 accuracy on YouTube spam detection. Zero Shot refers to prompting T0++ directly. +C means applying contextual calibration on the \bc{Voter} objects.}
    \label{tab:youtube}
\end{table}

In this experiment, we use Alfred to annotate the training split of YouTube spam detection dataset. \cite{alberto2015tubespam}
We replicate the setup used by \citeauthor{smith2022language}
The prompts are translated from the code-based labeling functions provided by the WRENCH benchmark \cite{zhang2021wrench}, a comprehensive weak supervision benchmark.
Alfred also includes a \bc{WrenchBenchmarkDataset} abstraction for easily running this benchmark.
In total, we define 10 prompted labeling functions with \bc{StringTemplate} objects.
Responses are mapped to votes using \bc{Voter} objects.
For this experiment, we use T0++ \cite{sanh2022multitask} as the backbone model for Alfred.
Following \citeauthor{smith2022language}, we also calibrate the responses from T0++ when voting using the contextual calibration strategy proposed by \citeauthor{zhao-2022-auto}.
Finally we aggregate the votes using the \bc{NaiveBayes} label model to produce the probablistic labels.
Table~\ref{tab:youtube} shows that Alfred makes reproducing the results of \citeauthor{smith2022language} easy, demonstrating that the combination of weak supervision and calibration yield a dramatic improvement over zero-shot prompting alone.

\subsection{Pet Breed Classification}
\begin{table}[h]
    \centering
    \begin{tabular}[width=0.5\textwidth]{cc}
    \hline
        Zero Shot & Prompted LFs \\
         \hline
         86.0  & 92.4 \\
         \hline
     \end{tabular}
    \caption{Top-1 accuracy on Oxford-IIIT Pet breed classification. Zero Shot refers to prompting CLIP directly.}
    \label{tab:pet}
\end{table}
Traditionally, programmatic weak supervision for vision has been limited by the ability to express supervision in code, relying on models such as object or attribute detectors to extract features and classify.
However, these object detectors often depend heavily on supervised training data, becoming a bottleneck for applying programmatic weak supervision in various vision tasks.
Fortunately, with large-pretrained vision-language model like CLIP \cite{radford2021learning}, we are now able to express supervision with natural language.
For this task, we develop prompts to classify 37 different breeds of pets from the Oxford-IIIT Pet dataset~\cite{parkhi2012cats}.
We use CLIP-ViT/L-14 as the backbone model and developed three simple prompted LFs.
The first two prompted LFs use templates ``a photo of [[label]]'' and ``a photo of [[label]] [cat/dog]'' where ``[cat/dog]'' is selected based on the breed.
The third prompted LF produces a partial label with the template "a photo of [cat/dog]", encouraging fine-grained labels to match with the coarse-grained type detected by CLIP.
We combine the votes using the \bc{NPLM} label \cite{yu2022nplm} to support weak supervision at various levels of granularity in the label space.
Table \ref{tab:pet} shows that this multi-granular weak supervision provides a nice boost over zero-shot prompting alone.

    \section{Discussion and Future Work}
This paper introduces Alfred, a prototype for the next generation of programmatic weak supervision systems that leverage the potential of large pre-trained models.
Alfred complements the existing ecosystem of large-pretrained-model toolkits, offering optimized inference throughput, a smooth local development experience, and compatibility with vision-language models to support image annotation tasks.
Alfred represents a notable advancement in the domain of programmatic weak supervision, as it enables users to express their domain-specific knowledge and heuristics with flexible natural language prompts for language and vision-language models. 
This approach can be more user-friendly than conventional PWS systems, which requires expert programming of weak supervision sources.
Our objective is for Alfred to serve as the infrastructure and experimentation platform for many future weak supervision research projects and applications. 
Furthermore, we plan to extend Alfred's capabilities to accommodate a wider range of multimodal large pre-trained models, such as Whisper \cite{radford2022robust} and LayoutLMs \cite{xu2020layoutlm, xu2020layoutlmv2, huang2022layoutlmv3}.
    
    \clearpage
    \section*{Limitations} 
Alfred is a prototype for the second generation of PWS systems, which incorporate large pre-trained models. 
However, there are some potential limitations to consider.
As with all PWS approaches, application quality is limited by the quality of the weak supervision sources used to vote on the labels.
In this case of prompted labeling functions, this depends on how well suited the prompts and model are to the task and domain.
If they are not well suited, then additional fine-tuning of the prompted models will be necessary.
Compared with traditional labeling functions written in code, understanding when and why labeling functions fail on certain examples can be particularly challenging.
Methods for explanations such as minimal contrastive edits~\cite{ross-etal-2021-explaining} can potentially help address this limitation.
We plan to explore incorporating such methods into Alfred.
    \section*{Ethics Statement}
One major concern for Alfred is the potential for biased or unfair labeling. 
Large pre-trained models are trained on massive datasets, which can reflect societal biases and inequalities. 
Consequently, supervision generated by these models can perpetuate and amplify these biases, leading to discrimination or unfair treatment in downstream applications. 
Therefore, it is essential to carefully consider the quality and representativeness of the backbone model for Alfred, as well as the prompts used for labeling data. 
To address potential labeling biases, human oversight and auditing are needed during the development loop to spot and correct any issues.
While Alfred has the potential to enhance the efficiency of programmatic data labeling, it is crucial to carefully consider and address potential ethical challenges.
    \section*{Acknowledgments}
We appreciate the helpful comments and discussion with Andrew Yuan, Avi Trost, Nihal Nayak and Zheng-Xin Yong.
This material is based on research sponsored by Defense Advanced Research
Projects Agency (DARPA) and Air Force Research Laboratory (AFRL) under agreement number
FA8750-19-2-1006. The U.S. Government is authorized to reproduce and distribute reprints for
Governmental purposes notwithstanding any copyright notation thereon. The views and conclusions
contained herein are those of the authors and should not be interpreted as necessarily representing
the official policies or endorsements, either expressed or implied, of Defense Advanced Research
Projects Agency (DARPA) and Air Force Research Laboratory (AFRL) or the U.S. Government. We
gratefully acknowledge support from Google and Cisco. Disclosure: Stephen Bach is an advisor to
Snorkel AI, a company that provides software and services for weakly supervised machine learning.
    
    \bibliography{main}
    \bibliographystyle{acl_natbib}

\end{document}